\documentclass[acmtog]{acmart}
\acmSubmissionID{385}

\usepackage{booktabs} % For formal tables
\usepackage{url}
\usepackage{multirow}
\usepackage{subcaption}
\usepackage[shortlabels]{enumitem}
\usepackage{subfiles}
\usepackage[export]{adjustbox}
\usepackage[super]{nth}
\usepackage{appendix}
\usepackage{sidecap}
\usepackage{soul}
\graphicspath{{figures/}{../figures/}}

\usepackage[ruled]{algorithm2e} % For algorithms

\SetAlFnt{\small}
\SetAlCapFnt{\small}
\SetAlCapNameFnt{\small}
\SetAlCapHSkip{0pt}

\newcommand{\etal}{\textit{et al}.}
\newcommand{\ie}{\textit{i}.\textit{e}.}
\newcommand{\eg}{\textit{e}.\textit{g}.}

\begin{document}

\title{Synthetic Defocus and Look-Ahead Autofocus for Casual Videography}

%% Authors.
%\author{Xuaner Zhang$^{1}$, Kevin Matzen$^{2}$,Vivien Nguyen$^{1}$,Dillon Yao$^{1}$,You Zhang$^{3}$,Ren Ng$^{1}$     }

%\affiliation{% \department{Department of Electrical Engineering and Computer Science}
%  \institution{$^{1}$University of California, Berkeley, $^{2}$Facebook Research, $^{3}$Chapman University}}
%\email{cecilia77@berkeley.edu}
%
\author{Xuaner Zhang}
\affiliation{\institution{University of California, Berkeley}}
\author{Kevin Matzen}
\affiliation{\institution{Facebook Research}}
%%\email{matzen@fb.com}
\author{Vivien Nguyen, Dillon Yao}
\affiliation{\institution{University of California, Berkeley}}
\author{You Zhang}
\affiliation{\institution{Chapman University, Independent Filmmaker}}
\author{Ren Ng}
\affiliation{\institution{University of California, Berkeley}}

% This command defines the author string for running heads.
\renewcommand{\shortauthors}{Zhang, Matzen, Nguyen, Yao, Zhang and Ng}

\authorsaddresses{}

% abstract
\begin{abstract}
In cinema, large camera lenses create beautiful shallow depth of field (DOF), but make focusing difficult and expensive. Accurate cinema focus usually relies on a script and a person to control focus in realtime. Casual videographers often crave cinematic focus, but fail to achieve it. We either sacrifice shallow DOF, as in smartphone videos; or we struggle to deliver accurate focus, as in videos from larger cameras. This paper is about a new approach in the pursuit of cinematic focus for casual videography. We present a system that synthetically renders refocusable video from a deep DOF video shot with a smartphone, and analyzes {\em future} video frames to deliver context-aware autofocus for the current frame. To create refocusable video, we extend recent machine learning methods designed for still photography, contributing a new dataset for machine training, a rendering model better suited to cinema focus, and a filtering solution for temporal coherence. To choose focus accurately for each frame, we demonstrate autofocus that looks at upcoming video frames and applies AI-assist modules such as motion, face, audio and saliency detection. We also show that autofocus benefits from machine learning and a large-scale video dataset with focus annotation, where we use our RVR-LAAF GUI to create this sizable dataset efficiently. We deliver, for example, a shallow DOF video where the autofocus transitions onto each person {\em before} she begins to speak. This is impossible for conventional camera autofocus because it would require seeing into the future.
\end{abstract}

\setcopyright{acmcopyright}
\acmJournal{TOG}
\acmYear{2019}\acmVolume{38}\acmNumber{4}\acmArticle{30}\acmMonth{7} \acmDOI{10.1145/3306346.3323015}

%CCS
%keywords

% A "teaser" figure, centered below the title and authors and above the body of the work.
\begin{teaserfigure}
  \centering
  \includegraphics[width=\linewidth]{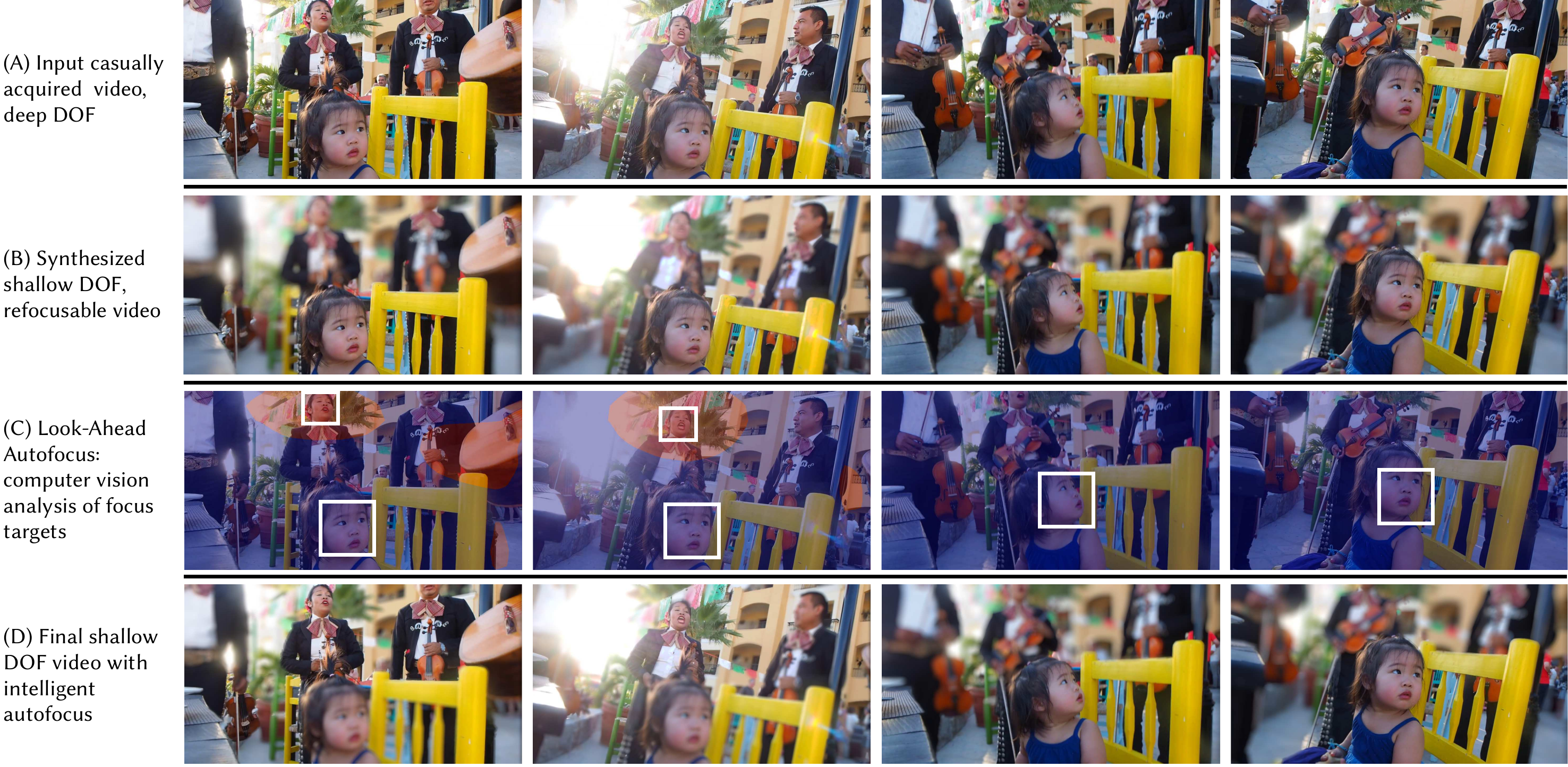}
  \caption{We present a new approach to pursue cinema-like focus in casual videography, with shallow depth of field (DOF) and accurate focus that isolates the subject. We start with (A) a deep DOF video shot with a small lens aperture. We use a new combination of machine learning, physically-based rendering, and temporal filtering to synthesize (B) a shallow DOF, refocusable video. We also present a novel Look-Ahead Autofocus (LAAF) framework that uses computer vision to (C) analyze upcoming video frames for focus targets. Here, for example, we see face detection (white boxes) and localization of who is speaking/singing~\cite{owens2018audio} (heat map). The result is shallow DOF video (D), where LAAF tracks focus on the singer to start, and transitions focus to the child as the camera pans away from the musicians. The LAAF framework makes future-aware decisions to drive focus tracking and transitions at each frame. This presents a new framework to solve the fundamental realtime limitations of camera-based video autofocus systems.}
  \label{fig:teaser}
\end{teaserfigure}

% Processes all of the front-end information and starts the body of the work.
\maketitle

\section{Introduction}
Cinematic focus is characterized by the beautiful, shallow depth of field (DOF) of large lenses, which are prized for their ability to visually isolate movie stars, control the viewer's gaze, blur out backgrounds and create gorgeous ``bokeh balls'' of defocused color. Focus that is tack sharp is essential, but shallow DOF makes it difficult and expensive to achieve. On a movie set, the primary camera assistant (``focus puller'') must operate the camera focus controls in realtime to track moving subjects and transition focus according to the screenplay. In movies with improvisational acting, such as Coherence (2014), the cinematographer must try to anticipate what the actors will do; of course significant focus error must be accepted. It is far more common to have a movie script, and for the focus puller to give actors markers on the ground to indicate where they should stand at specific points to facilitate highly accurate focusing. 

These are the issues that make cinematic focus impossible for casual videographers, even though we would love to achieve the aesthetic. Instead, with smartphone videography we sacrifice cinematic DOF, because the small lenses cause essentially everything to be in focus at the same time. In contrast, with cameras that have larger sensors and lenses capable of cinema-like DOF, we inevitably sacrifice focus accuracy. The reason is that there is no movie script, so memorable moments and decisive actions occur unpredictably. Like the focus puller for the Coherence movie, the camera's autofocus system would need a crystal ball to perfectly track each moving subject, or decide to transition focus to a new target in anticipation of its actions taking control of the narrative.

In this paper, we argue that a new direction is necessary if we are to truly deliver cinema-like focus for casual videography. An example unprecedented result we seek is a shallow DOF video of a group conversation where the focus transitions perfectly from person to person {\em before} each person begins talking. Another example is a video that faithfully tracks focus on a rapidly moving soccer player, and then presciently pulls focus onto another player before she heads the ball into the goal. The key to our new approach is to capturing refocusable video, and open the door to analyzing {\em future} video frames in order to deliver accurate tracking and anticipatory decisions about whether to transition focus to a decisive action by a new target. To enable these conventionally impossible capabilities, we contribute a framework composed of two modules.
\begin{enumerate}[wide]
\item
{\bf Refocusable Video Rendering (RVR)} Rather than capturing regular videos with static focus, we produce synthetic ``refocusable video'' where the focus depth of each frame can be computationally changed after capture. Our approach is to synthetically render a shallow DOF video, from a deep DOF video that can be recorded with a smartphone. We build on recent methods in this vein, which are limited to still photography and suffer from disturbing temporal inconsistency when applied frame by frame to videos. We extend synthetic shallow DOF to full video using a combination of machine learning, physically-based rendering and temporal filtering. For machine learning, we contribute a dataset of over 2,000 image pairs or triplets where the aperture and/or focus are varied, and use these data to train a convolutional neural network that improves prediction of RGBD video and recovery of HDR as input to Refocusable Video Rendering (RVR). 
\vspace{1mm}
\item {\bf Look-Ahead Autofocus (LAAF) for Casual Videography} We introduce the notion of Look-Ahead Autofocus that analyzes the seconds of video frames ahead of the current frame in order to decide whether to maintain or transition the focal depth. We demonstrate LAAF with examples of ``AI-assistance'' that include: motion and face detection to focus on upcoming human actions, audio localization to focus on who is about to speak, and a machine-learning-based focus detector that shows how a large-scale video dataset can be used to help autofocus of more generic videos. We build an interactive GUI incorporating with subject tracking and automatic focus transition so that the user only makes focus choices on a few keyframes to render a video with shallow DOF and annotated focus.
\end{enumerate}
\label{sec:intro}
\section{Prior Art and Related Work}
\subsection{Camera Autofocus Systems}
Camera autofocus systems have generally been classified into two buckets: contrast-detection autofocus (CDAF) and phase-detection (PDAF). CDAF is slower, seeking focus by aiming to maximize image contrast as the lens focus is changed; it performs poorly for video because the ``focus seeking'' behavior is visible in the recorded video. Phase detection can be much faster, and is based on separately detecting and comparing light passing through different parts of the lens aperture. This was achieved in SLR cameras by reflecting light onto PDAF units that each comprised a microlens atop multiple pixels~\cite{goldberg1992camera}.  

To enable PDAF in mirrorless camera designs, sensor makers began embedding microscopic PDAF units sparsely into the pixel arrays themselves~\cite{fontaine2017survey}, and advanced to the point that every imaging pixel became a PDAF unit to maximize light and autofocus-sensitive area~\cite{morimitsu20154m,kobayashi2016low}. This last design is now common in smartphones~\cite{fontaine2017survey,levoy2017portraitmode}.  One might argue that such advances in physical autofocus systems are asymptotically approaching the fastest possible in many devices today.  
And yet, autofocus mistakes remain common and inevitable in casual videography, because the focus of each frame is ``locked in" as it is being shot. A full solution is impossible because the autofocus algorithm would have to predict the future at every frame to correctly determine what to focus on, or transition focus to. 
This paper aims to lift this fundamental limitation by synthesizing shallow depth of field as a video postprocess, and using video autofocus algorithms that ``look ahead'' to make contextually meaningful predictions about what to focus on.

\subsection{Light Field Imaging}
Another way to capture refocusable images is a light field camera, but the cost of light field video systems remains very high. Rather than capturing a 2D slice of the light field, as is the case with a conventional camera that samples a set of rays that converge at a single point, commercial light field cameras, \eg, Lytro ILLUM\footnote{\url{https://www.dpreview.com/products/lytro/compacts/lytro_illum}} capture the full 4D slice of the light field. Light field imaging not only captures the set of rays from different viewpoints~\cite{levoy1996lightfield}, but also enables physically-accurate synthetic aperture rendering and after-the-fact refocusing~\cite{isaksen2000dynamically,Ng2005LightFP,wilburn2005high}. Beyond spatial resolution trade-offs, commercial light field video cameras currently have decreased video frame rate, approximately 3 FPS rather than the desired minimum of 30 FPS. Wang~\shortcite{wang2017light} propose a hybrid system using one light field camera and one DSLR camera to produce 30 FPS light field video view interpolation.

\begin{figure*}
  \centering
  \includegraphics[width=1\linewidth]{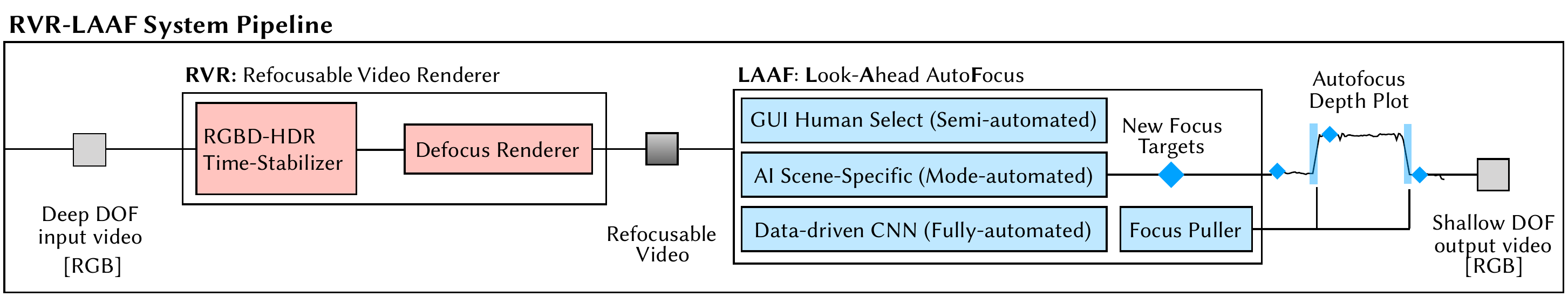}
\caption{
\label{fig:pipeline}
Overall system pipeline to compute shallow DOF video from a deep DOF video input. The Refocusable Video Renderer (RVR) contains a time-Stabilized RGBD-HDR (Section ~\ref{sec:rvr}) that computes temporal-stabilized depth and recovers HDR. The Look-Ahead Autofocus (LAAF) (Section ~\ref{sec:laaf}) pipeline consists of three approaches with different levels of automation, detecting New Focus Targets and generating autofocus depth followed by a focus puller that smooths focus transition. Output is a shallow DOF video with contextually-meaningful focus.  
}
\end{figure*}

\subsection{Synthetic Defocus}
Regarding synthesizing defocus in images, previous work has achieved breakthroughs on still photography and led to common ``portrait'' modes in current smart phone photography that synthetically blur the background in portrait photography. To the best of our knowledge, ours is the first to push such synthetic defocus from still photography to full videography. The first step towards RVR is to accomplish single image synthetic defocus, we give a high-level review on synthetic defocus for still images.

Stereo can be used to derive necessary depth~\cite{Yu2011DynamicDO,2014TRStereo}. The dual pixel sensors described earlier can be used to provide stereo views of the scene through right and left halves of the lens aperture. This has been used to estimate depth for synthesizing defocus blur for smartphone computational photography~\cite{wadhwa2018,levoy2017portraitmode}.

Data-driven machine learning approaches have also proven valuable in synthetic defocus tasks using single images, a lot of which are for driving scenarios using specialized datasets~\cite{godard2017unsupervised, kuznietsov2017semi, garg2016unsupervised}. MegaDepth~\cite{li2018megadepth} is a concurrent work that targets more generic image contents. Most related to our own work is the method of Srinivasan~\etal~\shortcite{srinivasan2017aperture}, who use weak supervision by predicting a depth map for an input photo, passing it through a differentiable forward rendering model, and then applying a reconstruction loss to the output shallow depth-of-field photo. However, they employ a simplified forward model and avoid having saturated pixels (e.g. salient bokeh regions) in their dataset. Park~\etal\shortcite{park2017unified} combines hand-crafted features with deep features to render refocusable images. However, they focus on noticeable defocus that is generated by medium-large aperture sizes. Therefore their method degrades on images taken by $\sim$f/16 and smaller aperture sizes, while this paper focuses on deep DOF input videos generated by f/20 or smaller.

A number of works also use defocus as a cue to predict depth of the scene~\cite{mather1996cue}. Nayar and Nakagawa~\shortcite{1994tpamishape} propose a focus operator which compares texture variability between images to determine relative level of focus. Correspondences from light-field imaging can be used to reduce ambiguity in depth-from-defocus~\cite{2013iccvtao}. Suwajanakorn~\etal~\shortcite{2015cvprSuwajanakorn} recently explored depth-from-defocus for smart-phone imagery and Tang~\etal~\shortcite{2017cvprtang} generalize depth-from-defocus for unconstrained smartphone imagery in the wild using two perceptibly similar images.

\subsection{Video Analysis}
Understanding video contents, such as knowing when and where activity happens or which regions are visually salient, is key to our video autofocus algorithm. Recent advances in video understanding tasks such as activity classification~\cite{soomro2012ucf101,karpathy2014large}, activity recognition and detection~\cite{sung2012unstructured,feichtenhofer2016convolutional,bilen2016dynamic}, benefit LAAF in localizing action in videos. Video saliency detects salient subjects under a more generic context, often making use of eye tracking signals~\cite{wang2018revisiting,sun2018sg}. We find saliency detection effective in proposing a coarse focus region for LAAF, and it can be combined with other detectors for finer-grained localization. A topic  similar to video attention along the temporal axis is video summarization~\cite{zhang2016video,mahasseni2017unsupervised}, extracting keyframes or subshot as visual summaries for long videos. More recently, audio-visual signals have been combined jointly to learn semantically meaningful video representations, one application we use to detect and locate people who are speaking is audio source localization~\cite{owens2018audio, ephrat2018looking}.

\label{sec:rel_work}
\section{System Overview: RVR-LAAF}
\begin{figure*}
  \centering
  \includegraphics[width=1\linewidth]{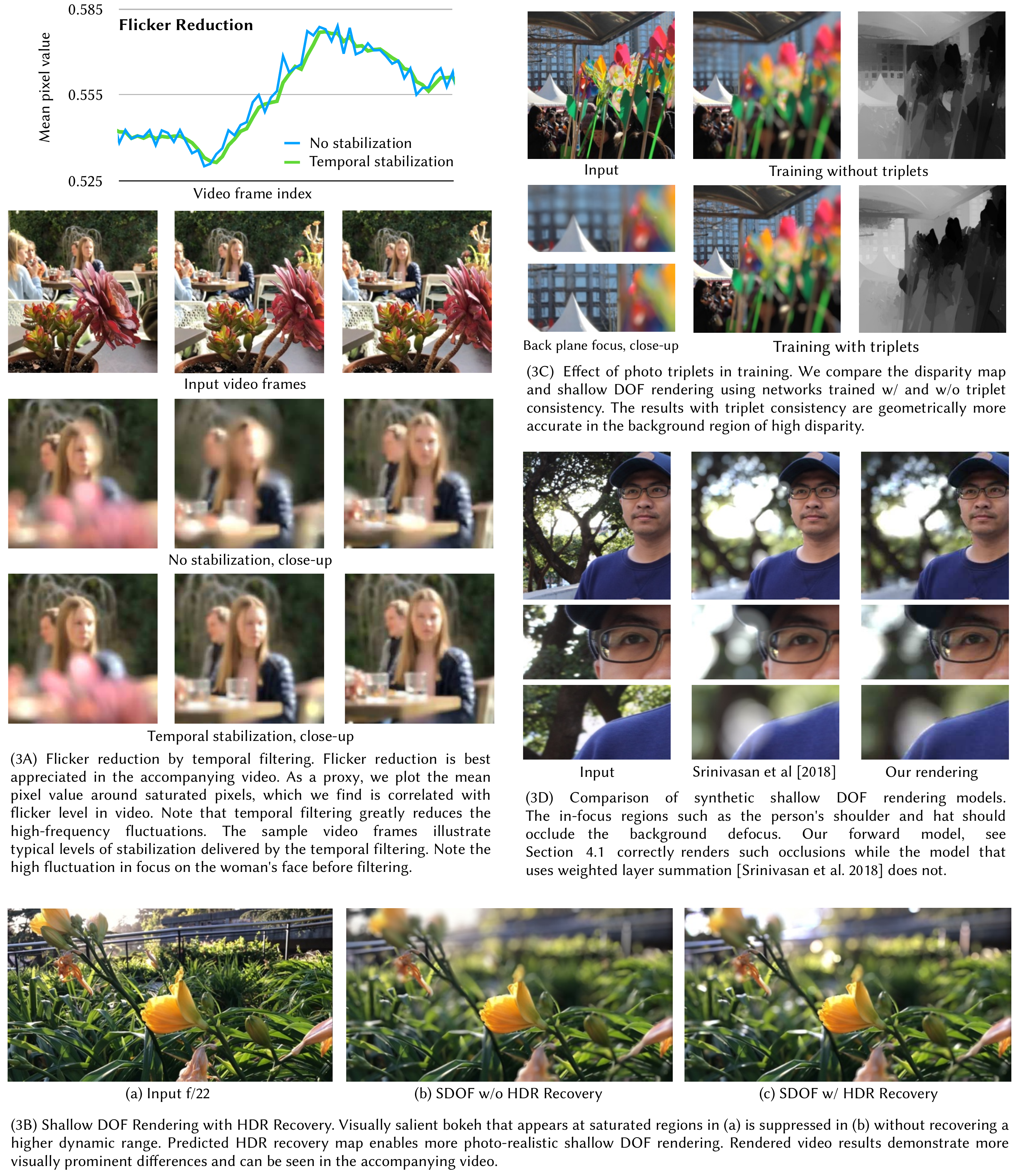}
\caption{
\label{fig:rvr-all}
Summarized contribution of our Refocusable Video Rendering subsystem (Section~\ref{sec:rvr}).}
\end{figure*}

\begin{figure*}
\includegraphics[width=1\linewidth]{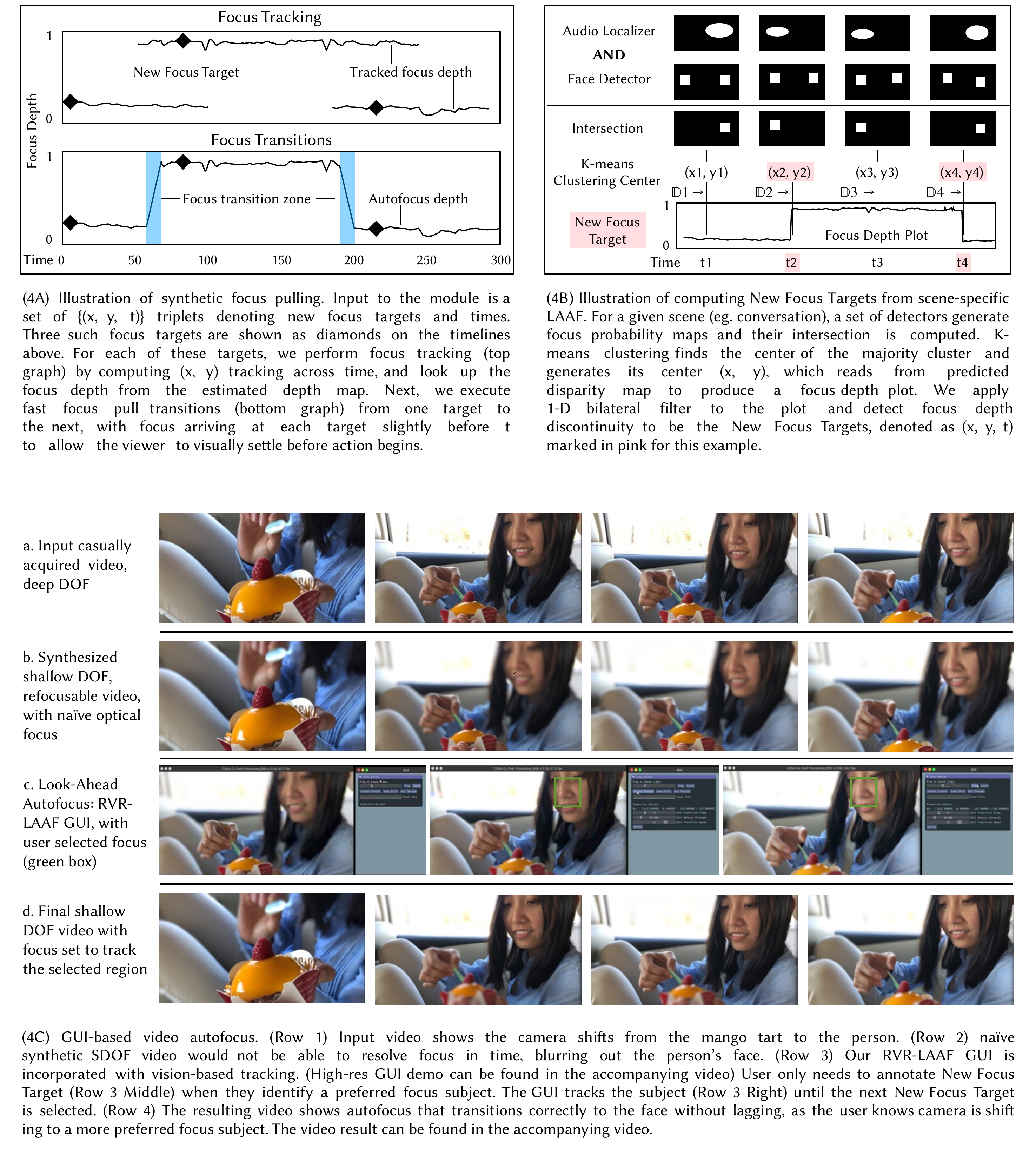}
\caption{Summarized contribution of our Look-Ahead AutoFocus subsystem (Section~\ref{sec:laaf}).}
\label{fig:laaf-all}
\end{figure*}

\renewcommand{\thefigure}{\arabic{figure} (Cont.)}
\addtocounter{figure}{-1}
\begin{figure*}
\includegraphics[width=1\linewidth]{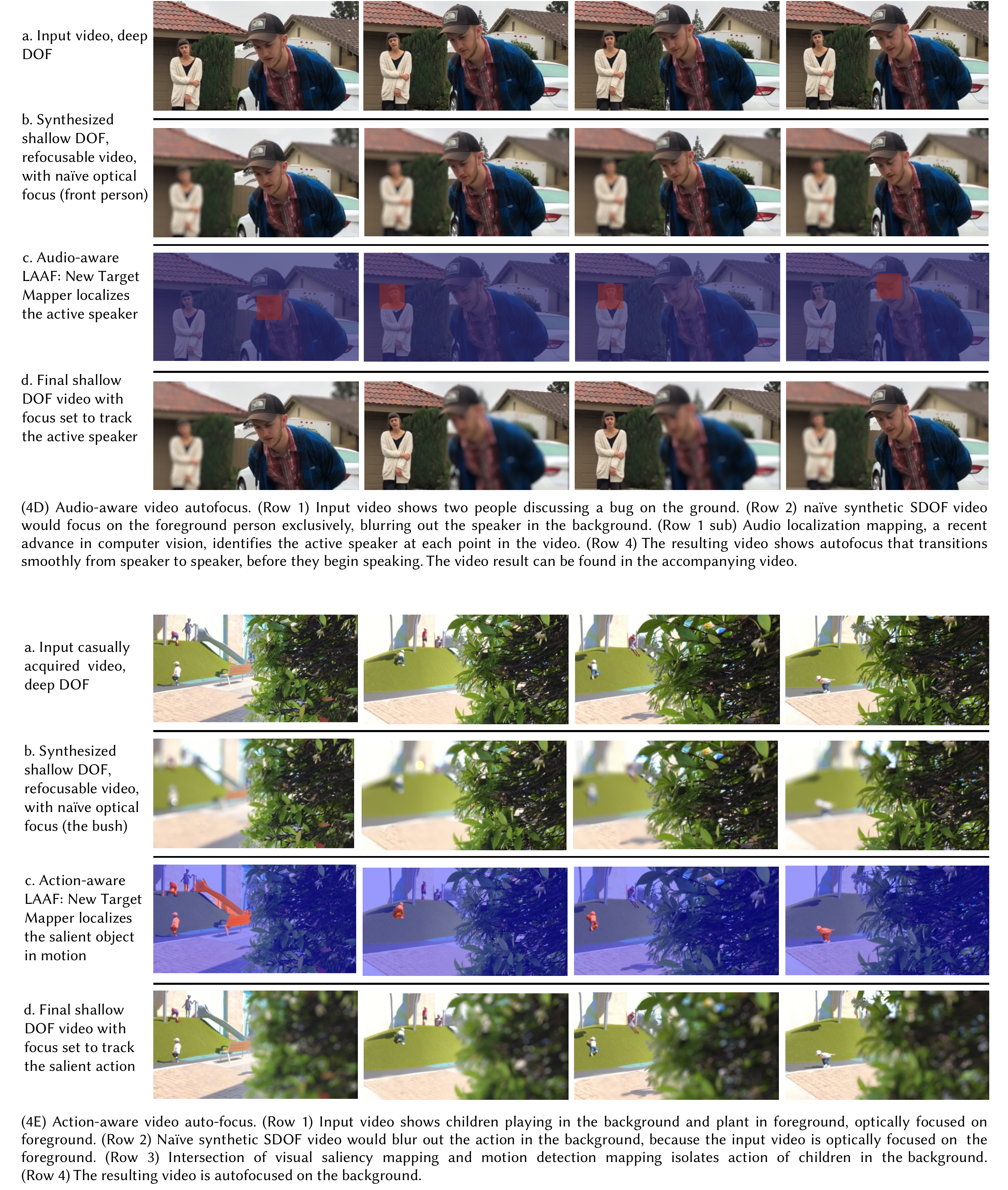}
\caption{Summarized contribution of our Look-Ahead AutoFocus subsystem (Section~\ref{sec:laaf}).}
\end{figure*}
\renewcommand{\thefigure}{\arabic{figure}}

\label{sec:overview}
Our system (See Figure~\ref{fig:pipeline}) aims to render cinematic autofocus for casual videos. It consists of two components: Refocusable Video Rendering (RVR) (Section~\ref{sec:rvr}) --- rendering videos with shallow DOF focused at any depth at any time, and video Look-ahead Autofocus (LAAF) (Section~\ref{sec:laaf}) --- choosing when and where to focus to make the autofocus choices contextually meaningful and visually appealing.

RVR is built upon a refocusable single frame renderer and a temporal module. We summarize our contribution in rendering refocusable video in Figure~\ref{fig:rvr-all}. We find it key to render RVR with temporal coherence, estimation of HDR detail, and a physically-based forward model of lens defocus. We achieve temporal stability by applying an occlusion-aware temporal filtering that is based on optical flow and robust to outliers (see Section~\ref{subsec:syn-temp} and Figure~\ref{fig:rvr-all}A). For photo-realistic rendering, we train a neural network to jointly estimate, from a single image, the defocus size and unclipped intensity value for each pixel. We find HDR recovery enables rendering of realistic bokeh (Figure~\ref{fig:rvr-all}B) and a correct forward model enables correct occlusion effects (Figure~\ref{fig:rvr-all}D). To train the network, we collect a large-scale aperture dataset that contains image pairs and triplets. We find our collected triplet dataset to improve estimation around large disparity regions (Figure~\ref{fig:rvr-all}C). We call our trained network a RGBD-HDR estimator (Section~\ref{subsec:syn-rgbd-hdr}). RVR takes a deep DOF video that can be captured by a smartphone, and generates a shallow DOF video that can be focused on any depth at any frame.

RVR delivers video that can be focused at any depth, but the question remains: what is the correct depth to focus on at every frame? For example, retaining optical focus and simply synthesizing shallow DOF (see Figure~\ref{fig:laaf-all}C row 2, and the video at 03:18) results in obvious focus errors and often lacks contextual meaning (e.g. focusing on one person while another person, blurred out, is speaking). 

Our solution is called the LAAF framework and comprises three complementary techniques for attacking this problem of ``when and where to focus''. First (Section~\ref{subsec:laaf-anno}), LAAF contains a carefully designed user interface (RVR-LAAF GUI) that enables a user to specify only a small number of semantically meaningful ``new focus targets'' in a video clip -- the system then tracks these subjects to maintain focus on them, and adds focus-pull transitions automatically. Second (Section~\ref{subsec:laaf-ai}), LAAF provides AI-based autofocus modules that only requires the user to choose the type of scene -- then, the system fully automates the task by choosing ``new focus targets'' intelligently. We demonstrate examples for scenes that contain conversations or actions, where the system automatically pre-focuses on each person before they speak or on actions before they occur. Third (Section~\ref{subsec:laaf-generic-vid}), (Section 5.3), we demonstrate a first attempt at fully automating video autofocus using a machine learning approach -- we contribute ground-truth focus annotations on a large-scale video dataset, using our RVR-LAAF GUI to create this sizable dataset efficiently.

\label{sec:oveview}
\section{Refocusable Video Rendering (RVR)}
In this paper we introduce the first synthetic defocus renderer for videos. We take Monocular Depth Estimation (MDE)~\cite{srinivasan2017aperture} as our launching point, and comprehensively re-work the method to make it suitable for synthesizing defocus for casual videography. This section presents the implementation details. The four critical changes are:
\vspace{-1mm}
\begin{enumerate}[wide]
    \item Adding HDR recovery estimation, which we show is critical to achieving plausible bokeh balls that are a visual hallmark of shallow DOF videos (Figure~\ref{fig:rvr-all}B, Section~\ref{subsec:syn-rgbd-hdr})
    \item Building a superior training dataset with far greater scene diversity. This dataset contains novel "triplets" of images (Figure~\ref{fig:dataset}) that we find is important to improve depth estimation in foreground and background regions (Figure~\ref{fig:rvr-all}C Section~\ref{subsec:syn-rgbd-hdr}).
    \item Correcting the forward model to make it handle occlusions correctly and so that background defocus does not incorrectly "bleed" around foreground objects (Figure~\ref{fig:rvr-all}D, Section~\ref{subsec:syn-rgbd-hdr})
    \item Adding temporal coherence to ameliorate flicker (Figure~\ref{fig:rvr-all}A, Section~\ref{subsec:syn-temp})
\end{enumerate}

\subsection{RGBD-HDR Estimator}
\label{subsec:syn-rgbd-hdr}
We offer an indirectly-supervised approach to estimate disparity and HDR with only aperture supervision, by training a neural network that jointly predicts a disparity map $\mathbb{D}$ and recover high dynamic range (HDR) $\mathbb{E}$ from a single image that has a deep DOF and standard dynamic range (SDR). $\mathbb{D}$ and $\mathbb{E}$ will be formally defined later. Unlike the MDE dataset that avoids saturation problems by avoiding saturated pixels in the input photo, our extended MDE dataset has a diverse set of scenes that cover a wide dynamic range and contain saturated pixels that are common in casual videography. We find that both disparity and HDR can be estimated purely from our extended MDE dataset, by imposing supervision on the reconstructed shallow DOF images. Alternative to our joint prediction is to estimate depth and HDR separately in sequence, for example, an HDR recovery network followed by a depth prediction network. Recent works have addressed performance on monocular depth estimation~\cite{li2018megadepth} and monocular HDR recovery~\cite{EKDMU17}. Our sub-system on monocular shallow DOF rendering would be approximately equivalent to a composition of these state-of-the-art works. One of our advantages is that we do not require ground truth supervision on either depth or HDR, while the aforementioned methods use direct supervision and thus require challenging data capture and annotation to account for model generalization to different or more generic scene content.

We formulate our joint disparity and HDR estimation network as following. Given training dataset $\mathcal{A}$ with pairs of small aperture input $I^{S}$ and large aperture ground truth output $I^{L}$, the major loss we apply to optimize the network parameters is the rendering loss:
\begin{equation}
\begin{aligned}
L_{\mathrm{rend}} = \|I^{L} - \mathcal{F}(I^{S}, \mathbb{D}, \mathbb{E})\|_1
\end{aligned}
\end{equation}
We use shallow DOF rendering as the objective, thereby bypassing the need to have direct supervision using disparity or HDR ground truth, which can be extremely challenging to collect and annotate. 

The forward model $\mathcal{F}$ is based on an ideal thin lens model~\cite{potmesil1982synthetic}. It takes a deep DOF image with predicted $\mathbb{D}$ and $\mathbb{E}$ to render a synthetic shallow DOF image. We use a disk kernel $K$ to approximate the point spread function of a defocused point through the lens. To handle occlusion, we blend layers of different disparity levels in order from back to front to prevent background blur from incorrectly bleeding around the silhouette of foreground objects, as shown in Figure~\ref{fig:rvr-all}D. Previous methods such as~\cite{srinivasan2017aperture} simply sum up all disparity levels.

We define disparity $\mathbb{D}$, the inverse depth, in its stereo sense being created by the differing vantage points from left and right edges of the lens aperture. This amount is proportional to the defocus blur size in pixel space by a normalized scalar~\cite{held2010using}. Assume the disparity map $\mathbb{D}$ ranges from $d_{\mathrm{min}}$ to $d_{\mathrm{max}}$ and denote disk kernel radius as $r$ ($r=0$ at focal plane). We discretize $\mathbb{D}$ into $|d_{\mathrm{max}} - d_{\mathrm{min}}|$ levels using a soft mask $M$. We define $M$ as the matte for content at the corresponding disparity level. $M$ allows the forward model $\mathcal{F}$ to be differentiable and also stabilizes training. Formally, $M$ at disparity $d$ is defined as:
\begin{equation}
\begin{aligned}
M(\mathbb{D},d) = \exp(-\lambda(\mathbb{D}-d)^2)
\end{aligned}
\label{eq:obj}
\end{equation}
$\lambda$ is empirically set to $2.8$ to model a continuous and rapid falloff across neighboring disparity levels. Because we predict signed disparity and define focal plane to have zero disparity, $r=d (d \geq 0)$ when $d$ is behind the focal plane, otherwise, $r=-d (d < 0)$. We use $I^{l}$ to denote the rendered shallow DOF image. Using our back-to-front forward rendering model, the shallow DOF image at disparity level $d$, denoted as $I_{d}^{l}$, is computed by blending it with its previous disparity level $I_{d-1}^{l}$ using the aforementioned content mask $M$:
\begin{equation}
\begin{aligned}
I_{d}^{l}= I_{d-1}^{l} \cdot (1-M(\mathbb{D},d)) + \big( I^{S} \cdot M(\mathbb{D},d) \big) \otimes K(r)\\
\end{aligned}
\label{eq:forward}
\end{equation}

\begin{figure}[h]
  \includegraphics[width=\linewidth]{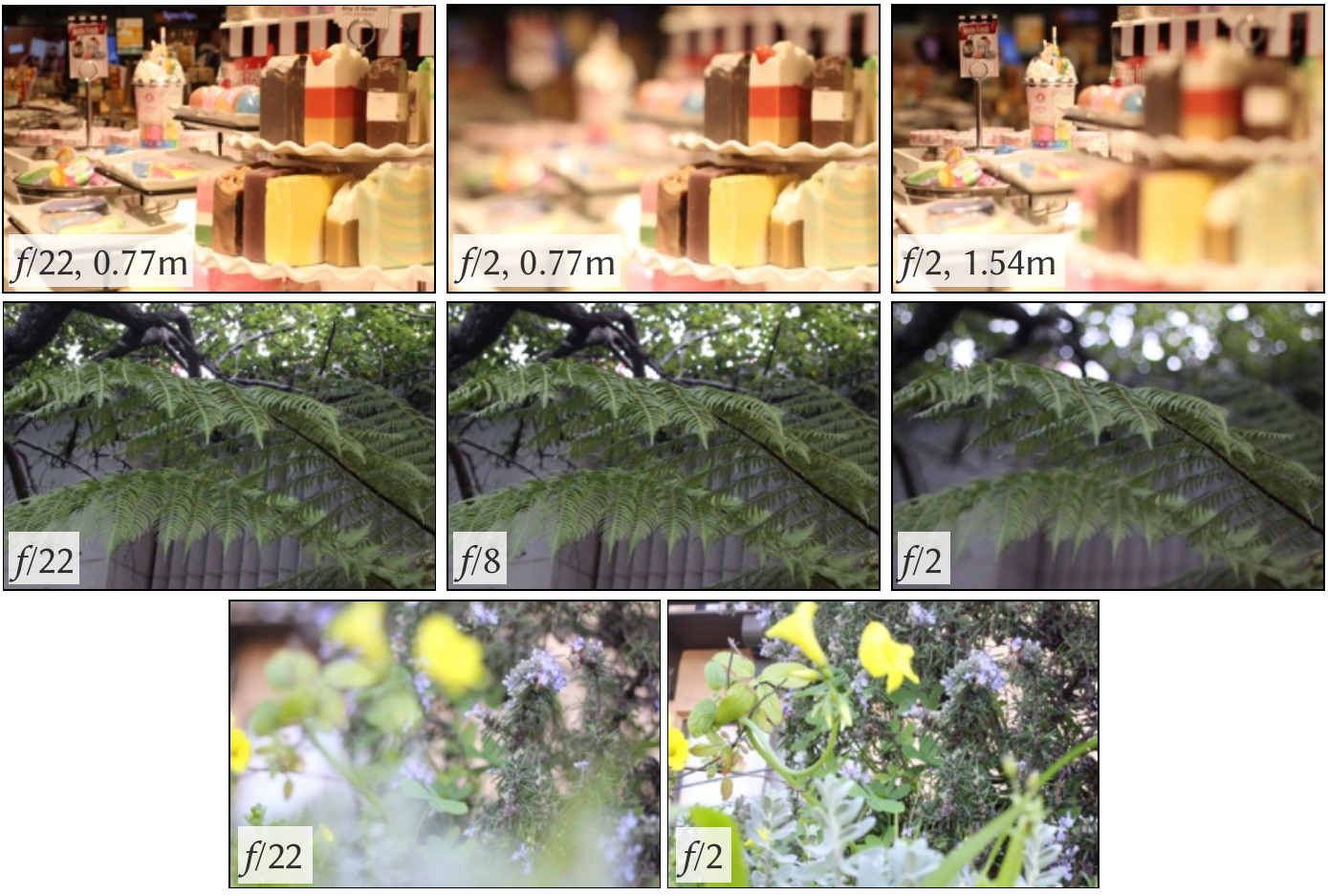}
  \caption{Example image pairs and triplets in our dataset. (Row 1) A focal triplet example, defocus map predicted from the input image is used to reconstruct the image with the same focus depth but taken with a large aperture (middle image) shifted to reconstruct the image taken with a large aperture at a different depth plane (right image). (Row 2) An example of aperture triplet, defocus map predicted from the input image is used to reconstruct the large aperture image, and scaled to reconstruct the medium aperture image. (Row 3) An example of image pairs, note that the focus is on the middle plane for this example.}
\label{fig:dataset}
\vspace{-5mm}
\end{figure}

One of the beautiful and characteristic visual signatures of shallow DOF video are so-called "bokeh balls," which are bright disks optically created by defocusing of small, bright lights (see the background of the middle image in Figure~\ref{fig:dataset} and 02:03 in the video). A numerical challenge for synthesizing such bokeh balls is that one needs HDR images to capture the very high intensity of the small lights~\cite{debevec2008recovering}. Without HDR, the represented intensity of the small bright lights is limited by the 8-bit fixed point precision of our input videos  -- when synthetically defocusing these lights, the incorrectly low intensity value spreads out over many pixels and becomes invisible rather than a bright ball of light (see middle image in Figure~\ref{fig:rvr-all}B). Therefore, we find HDR recovery key to render visually salient bokeh that appears at saturated regions, which has not been considered in prior synthetic defocus rendering models~\cite{kraus2007depth, Yang2016VirtualDH, wadhwa2018}. We undo gamma correction on input images to work in linear space. We predict $\mathbb{E}$ in log scale to recover a high dynamic range. Pixels that are saturated in the deep DOF image are often not saturated in its shallow DOF pair, because their energy is spread over many pixels. This provides indirect signal for the network to learn HDR recovery. We replace $I^{S}$ in Equation~\ref{eq:forward} with its HDR version $I^{S'}$, computed as:
\begin{equation}
\begin{aligned}
I^{S'} = I^{S} \cdot e^{k \cdot \mathbb{E}}
\end{aligned}
\label{eq:hdr}
\end{equation}
where $k$ affects the maximum recovered saturation value and is empirically set to 50.

\paragraph{Data Collection.}
\label{subsec:syn-data-collection}
To train our RGBD-HDR Estimator, we build upon the Flower dataset~\cite{srinivasan2017aperture} and contribute the first large-scale aperture dataset that covers diverse object categories. The dataset contains 1.2K image pairs and 0.8K image triplets taken with different aperture sizes (f/2, f/8 and f/22) and focus depth. Each image pair or triplet is taken in a scripted continuous shot using Magic Lantern\footnote{\url{https://magiclantern.fm/}} firmware add-on for Canon EOS DSLR cameras. This minimizes misalignment among pairs/triplets during capturing time. Pixel-wise alignment is further imposed via correlation coefficient minimization~\cite{evangelidis2008alignment}.

\paragraph{Image Pair and Triplet Supervision.}
\label{subsec:syn-supervise}
The network takes in a deep DOF image and predicts a disparity map and a high dynamic range, which are used to render a shallow DOF image, and has ground truth to compare against. This rendering loss is back-propagated to update network parameters until convergence. We notice that the precision of large defocus values in $\mathcal{D}$ is less accurate as the gradient of the reconstruction loss decreases inversely proportional to the size of the defocus disk kernel ($\frac{\partial{L_{\mathrm{rend}}}}{\partial{r}}\propto\frac{1}{r^2}$). This produces visual artifacts when refocusing the video to planes that are originally at large disparity. To mitigate imbalanced loss gradient back-propagated through different disparity planes, we apply a triplet consistency checking during training. Our dataset contains two types of image triplets: aperture and focal triplets. \textit{Aperture triplets} are taken with f/2, f/8 and f/22. The estimated disparity should be able to \textit{scale} to render both median DOF and shallow DOF images. This constraint also helps stabilize training. In contrast, \textit{focal triplets} include a deep DOF image and two shallow DOF images focused at different depths. The estimated disparity map should be able to \textit{shift} to render both shallow DOF images at different depths. From our thin lens model assumption, the non-linearity between change of focal plane and change of defocus blur size only depends on the object disparity and lens movement in sensor coordinate, which is relatively small to be negligible. As long as the object is not too close to the camera, we can assume the shift of focal plane to be linear to the change of defocus size. For the prime lens, Canon EF 50mm f/1.8, that we use for data capture, scene depths difference from infinity to 0.5m generate a deviation of $\sim$5\% from the assumed linear model. Training with both types of triplet data helps to improve the precision of large disparity region estimation. Figure~\ref{fig:rvr-all}C shows a visual comparison on the estimated disparity map and back-focus shallow DOF rendering between training with and without triplet consistency data.

\paragraph{Loss Functions.}
We train RGBD-HDR with rendering objectives that penalize difference between rendered results and ground truth shallow DOF targets. We supervise the network with per-pixel $L^1$ loss, denoted as $L_{\mathrm{pix}}$, as well as low-level and high-level image features denoted as $L_{\mathrm{feat}}$, by feeding the network output and target through a pre-trained VGG-19 network $\Phi$. We compute the $L^1$ difference between $\Phi(\mathcal{F}(I^{S}, \mathbb{D}, \mathbb{E}))$ and $\Phi(I^{L})$ in selected feature layers.
\begin{equation}
\begin{aligned}
\label{eq:loss_rec}
L_{\mathrm{rend}}(I^{S}, I^{L})&=L_{\mathrm{feat}}(I^{S}, I^{L})+L_{\mathrm{pix}}(I^{S}, I^{L})\\
&=\sum_i{\lambda_i\|\Phi_i(I^{L})-\Phi_i(\mathcal{F}(I^{S}, \mathbb{D}, \mathbb{E}))\|_1}+\\
&\|I^{L}-\mathcal{F}(I^{S}, \mathbb{D}, \mathbb{E})\|_1,
\end{aligned}
\end{equation}
where $\Phi_i$ indicates the layer $i$ in the VGG-19 network. The weights $\{\lambda_i\}$ are used to balance different terms in the loss function. We select the layers \texttt{conv1\_2}, \texttt{conv2\_2}, \texttt{conv3\_2}, \texttt{conv4\_2}, and \texttt{conv5\_2} in the VGG-19 network. These features are demonstrated to be efffective for image enhancement, style transfer and many other image processing tasks~\cite{chen2017photographic,Johnson2016,xzhang2018reflectionremoval}.

Image triplets are applied in an adjusted rendering objective that penalizes difference from shallow DOF image $I^{L'}$ after adjusting aperture (scale) and focal plane (shift), with the adjustments linearly approximated by an affine model. Similar to $L_{\mathrm{rend}}$, $L_{\mathrm{adjust\_rend}}$ computes the $L^1$ difference between $\Phi(\mathcal{F}(I^{S}, \alpha\mathbb{D}+\beta, \mathbb{E}))$ and $\Phi(I^{L})$ in the same selected feature layers. We find this image triplet training effectively improves RGBD-HDR prediction at far planes, as shown in Figure~\ref{fig:rvr-all}C.

We additionally incorporate an edge-aware smoothness penalty $L_{\mathrm{smooth}}$ on the predicted disparity map by minimizing the $L^1$ norm of its gradients, weighted less on the input image edges. This ensures the predicted disparity map to be locally smooth and is formulated as:
$$L_{\mathrm{smooth}}(\mathbb{D}) = \|\partial_x\mathbb{D}\|_1 \cdot e^{-|\partial_x{I^{S}}|} + \|\partial_y\mathbb{D}\|_1 \cdot e^{-|\partial_y{I^{S}}|}$$
Overall, we train our network by minimizing a loss function that is a weighted sum of $L_{\mathrm{rend}}$, $L_{\mathrm{adjust\_rend}}$, and $L_{\mathrm{smooth}}$.
\begin{equation}
\begin{aligned}
\label{eq:loss}
L_{\mathrm{total}} = & \sum_{(I^{S},I^{L},I^{L'})\in \mathcal{A}} L_{\mathrm{rend}}(I^{S}, I^{L}) + L_{\mathrm{adjust\_rend}}(I^{S}, I^{L'})\\
& + w_1 L_{\mathrm{smooth}}(\mathbb{D})
\end{aligned}
\end{equation}
where $w_1$ is the weight for the smoothness regularization, and is set to 10 across all experiments. When a triplet is not available for a particular example, \ie, only a pair is available, we omit $L_{\mathrm{adjust\_rec}}(I^{S}, I^{L'})$ and double the weight of $L_{\mathrm{rend}}(I^{S}, I^{L})$.

\paragraph{Training and Implementation.}
We use a U-net network architecture~\cite{ronneberger2015u} that contains an encoder-decoder structure with skip connections. All layers are followed by a leaky \emph{ReLU} activation, except for the last prediction layer that produces $3+N$ channels. Three of these channels are used for HDR recovery. The other N channels are used as a bilateral-space representation over luma, which are sliced with a bilateral slicing operator (where \textit{N} is defined by bandwidth parameters of the bilateral slicing operator) into pixel-space to produce a 1-channel signed disparity map. In practice, we find that predicting a bilateral-space representation improves fidelity of the disparity map over predicting directly in pixel-space, particularly around edges. This is consistent with the findings of Gharbi~\etal~\shortcite{gharbi2017deep} and Barron~\etal~\shortcite{barron2015fast}. Positive disparity refers to planes behind the focal plane while negative disparity represents planes in front of the focal plane.

We train the network with batch size 1 on an NVIDIA Titan X GPU and weights are updated using the Adam optimizer \cite{KingmaBa2015} with a fixed learning rate of $10^{-4}$. A full network architecture will be made available in a code release. The network converges after $150K$ iterations. Our network is fully convolutional and can run at arbitrary image sizes. During training, we resize the images to random resolutions between $512$p and $1024$p.

\subsection{Video Temporal Consistency.}
Visually, we find that the most important change when going from still images to video is to enforce temporal consistency. Independently rendering each frame with shallow DOF causes visually disturbing flickering, especially around prominent bokeh. A comparison can be found in Figure~\ref{fig:rvr-all}A and 01:44 in the video. To impose temporal coherency, we apply a weighted temporal moving average that is occlusion-aware and robust to outliers to $\mathbb{D}$ and $\mathbb{E}$. For each target frame $I_i$, we compute $w_i,w_{i-1},...,w_{i-M}\in W$, where $M$ is the number of neighboring frames. We compute optical flow using a pre-trained deep neural network Flownet 2.0~\cite{ilg2017flownet} for consecutive pairs of frames, and align $I_{i-1},...,I_{i-M}$ to $I_i$ using concatenated flows. $w_i$ is computed as a weighted combination of an occlusion weight $w^{\mathrm{occl}}_{i}$ and an outlier weight $w^{\mathrm{med}}_{i}$.

\paragraph{Occlusion Weight}
Occluded pixels should be weighted little. We adopt the tactic of forward-backward consistency checking~\cite{ogale2005motion,chen2016full}, computing both forward $f^{i\rightarrow i+1}$ and backward optical flow $f^{i+1\rightarrow i}$ for frame pair $I_i$ and $I_{i+1}$. Consider point $p$ in $f_i$ that shifts to $p+f^{i\rightarrow i+1}(p)$. We check if we can find a point $q$ in $f_{i+1}$ such that:
\begin{equation}
\begin{aligned}
\|(p-f^{i+1\rightarrow i}(q))\|_2 + \|(q - p+f^{i\rightarrow i+1}(p))\|_2 \leq \delta
\end{aligned}
\end{equation}
where $\delta$ is a small distance threshold. If there exists such a $q$ in $f_{i+1}$, $p$ is considered as a consistent pixel in $f_i$. For each frame $I_{i}$ we compute such an occlusion mask and call it $w^{\mathrm{occl}}_{i}$.

\paragraph{Outlier Rejection}
To account for optical flow inaccuracies, we classify a warped pixel as an outlier if it has very different values within its temporal moving window. We assume that outliers are sparse and thus a majority vote approach such as median filtering would be effective. For any point $p$ in frame $I_i$, its outlier weight $w(p)_{i}$ is set to $e^{-\mathrm{median}(|p_i + f^{j\rightarrow i}(p) - p_{i}|_{j=i-M}^{i})}$. We compute an outlier weight for each frame and call it $w^{\mathrm{med}}_{i}$.
Overall, the filtered prediction $\mathbb{D}_{i}$ (or $\mathbb{E}_{i}$) of target frame $I_i$ is computed as:
\begin{equation}
\begin{aligned}
\mathbb{D}_{i} = \sum_{j=i-M,...,i-1}W_j \cdot (\mathbb{D}_i + f^{j\rightarrow i}) + W_i \cdot \mathbb{D}_i
\end{aligned}
\end{equation}
where $\sum_{i-M}^{i}W_j = 1$. We set $M$ to 6 for all experiments.
\subsection{Parallelization in \textit{Scanner}.}
\label{subsec:syn-temp}
We used a set of 100 test videos in developing and validating our prototype, so we use modern infrastructure to process them with RVR. We choose to use Scanner~\cite{poms2018scanner}, which gives us the option to process on different hardware and parallelize on the cloud. With full parallelism we could in principle process all 100 videos in under a minute. In practice we use a local 4-core machine with a single Titan X GPU. On average, we process one 2 megapixel video frame in ten seconds, including RGBD-HDR inference, bi-directional optical flow computation, and temporal filtering. The current bottleneck is the flow-based temporal filtering and the fact that we do not optimize the total processing time to pursue full advantage of Scanner's distributed processing of jobs.

\label{sec:rvr}
\section{Look-Ahead Autofocus (LAAF)}
Being able to synthesize refocusable videos is not a complete solution to generating a meaningful shallow DOF video -- deciding on when and where to focus in the video is challenging and essential. In the conversation example shown in Figure~\ref{fig:laaf-all}D, the focus should shift between the active speaker and lock onto the person about to speak next, \textit{before} she speaks. On a movie set, exact focus is achieved by a movie script that exhaustively defines what should be in focus at every moment in the film, and a dedicated focus puller (the \nth{1} Assistant Camera) who measures and marks the exact focus position. In this section, we demonstrate how LAAF uses recent computer vision advances in video understanding --- analyzing semantics of current and future frames --- to enable video autofocus that automates portions of focusing process in cinematography. Our prototypes demonstrate, we believe for the first time, that it is actually tractable to deliver. We demonstrate three approaches towards semi-automated and fully-automated video autofocus, using:

\begin{enumerate}[wide]
    \item an interactive GUI with vision-based tracking and simple human focus selection (Section~\ref{subsec:laaf-anno})
    \item scene-specific (\eg~conversation, action, etc.) AI-assist modules (Section~\ref{subsec:laaf-ai})
    \item a data-driven CNN network trained from a large-scale video dataset with focus annotation, labeled using our GUI (Section~\ref{subsec:laaf-generic-vid})
\end{enumerate}

Output from the above three approaches is a set of $\{(x,y,\hat{t})_i\}$ triplets denoting New Focus Targets --- focus regions and times. Three such New Focus Targets are shown as diamonds on the timelines in Figure~\ref{fig:laaf-all}A. For each New Focus Target, we perform focus tracking by computing $(x,y)$ tracking across time, and look up the focus depth or subject from the estimated depth map. Next, we execute a digital focus pull from one target to the next, with focus arriving at each target slightly before the target time, to allow the viewer to visually settle before action begins. We set a default duration of 10 frames ($\sim$ 0.67 sec) for a focus pull and linearly interpolate focal planes in between.

For real systems, LAAF could happen during video capture as well. To do this, we would need to buffer a few seconds of video frames, which would be the temporal window we set to look ahead, and then pipeline the LAAF processing with video recording.
\subsection{GUI-based Video Semi-Autofocus \\
--- Human Selection of New Focus Targets}
\label{subsec:laaf-anno}
We build an interactive RVR-LAAF GUI incorporating a vision-based tracker (\eg ~KCF tracker~\cite{henriques2015high}) such that the user only needs to specify a New Focus Target, instead of selecting a focus subject for each frame, which is extremely inefficient and impractical. The tracker follows the selected focus region until the user pauses the video to select the next New Focus Target (see 03:08 in the video).

One interesting point is that RVR-LAAF GUI provides benefit even for simple scenes that seem amenable to conventional autofocus. One might think that in these situations simply synthesizing shallow DOF from the recorded video would suffice. However, the issue is that synthetic defocus will amplify any misfocus error. For example, a subtle defocus from a person's eyes to back of the head would be almost unnoticeable in a deep DOF input image, but after synthetic defocus the person's features would become unacceptably blurred out -- in this situation LAAF can keep the focus locked maintain sharp focus on the person's features throughout the video. So even in these situations, RVR-LAAF GUI allows us to increase the focus accuracy of the output video by adjusting the synthetic focal plane onto the person of interest and track the corrected subject. In other words, autofocus achieved by LAAF is essential to delivering even simple synthetic defocus video accurately.

\subsection{Scene-Specific Video Autofocus \\
--- AI-Based Selection of New Focus Targets}
\label{subsec:laaf-ai}
Fully-automated video autofocus without human interaction requires visual and semantic understanding of the video context. We show how LAAF incorporates recent advances in video understanding to automate, to some extent, human choices of focus selections. Faces, actions, audio sources and other salient (visually distinctive) objects, are some common subjects to set in focus in casual videography. We exploit a set of context-aware detectors $\mathcal{H}$ (\eg~face, action detectors) to automate the generation of New Focus Targets. 

As illustrated in Figure~\ref{fig:laaf-all}B, we compute the intersection of a selected set of detection to identify a scene-dependent and contextually-meaningful focus region. We then apply K-means clustering (K empirically set to 4) to determine the majority clustering centroid position $(x,y)$ and read in the focus depth from the predicted disparity map $\mathbb{D}$. Frames with empty intersection use its previous disparity level. Next, we apply a bilateral filter followed by an edge detection to identify focus depth discontinuity, which marks the New Focus Target, $(x,y,t)$. Note that we use $(x,y)$ instead of $\mathbb{D}(x,y)$ to denote New Focus Target because we later use $(x,y)$ to track the subject to synthesize focus puller as shown in Figure~\ref{fig:laaf-all}A.

For example, we demonstrate scene-specific LAAF on two types of videos, one on action videos that use action and saliency detection ($\mathcal{H}=\{H_{\mathrm{act}}, H_{\mathrm{sal}}\}$) to detect salient regions that also involve action (Section~\ref{subsec:laaf-action-vid}); one on conversation videos that use audio localization and face detector ($\mathcal{H}=\{H_{\mathrm{aud}}\, H_{\mathrm{face}}\}$ to detect and focus on the person who is speaking (Section~\ref{subsec:laaf-conversation-vid}). These two types of videos are commonly seen in both casual videography and professional filmmaking. According to a video essay that attempts to compile a list of significant examples of focus racking in film history(years 1963-2016)~\footnote{\url{https://www.youtube.com/watch?v=tT_qv9ptauU&t=75s}}, half of the focus racks are triggered by human action or audio source change.

\paragraph{Action-aware LAAF}
\label{subsec:laaf-action-vid}
\noindent We collect a set of videos that involve unexpected actions, which trigger focus depth change towards the action subject. An example scenario is a background subject of interest entering the frame unexpectedly while focus is at the foreground, and the result we seek is to shift focus at, or a few frames before, the subject entering the view. We use an action localizer ($H_{\mathrm{act}}$) and a salient object detector ($H_{\mathrm{sal}}$) to generate probable focus regions that are both salient and involve actions. $H_{\mathrm{act}}$ is based on optical flow and deepmatch~\cite{revaud2016deepmatching} for action localization and tracking, and is used as video pre-processing step for several computer vision tasks such as unsupervised video feature learning~\cite{pathakCVPR17learning}. $H_{\mathrm{sal}}$ is based on a still image saliency detection~\cite{hou2017deeply} work, which trains a deep network to compute a saliency heat map that identifies visually distinctive objects and regions in an image. As illustrated in Figure~\ref{fig:laaf-all}E and 05:22 in the video, LAAF analyzes future frames (Row 1) and detects the child's action that is about to happen. Instead of always keeping focus on the foreground bush, LAAF is able to shift focus to the background before the child slides down the hill (Row 4).

\paragraph{Audio-aware LAAF}
\label{subsec:laaf-conversation-vid}
We demonstrate LAAF applied to another video collection of conversational scenes. The goal is to place focus right before the person who is about to speak. We use an audio localizer ($H_{\mathrm{aud}}$) and face detector ($H_{\mathrm{face}}$) to compute probable focus regions of the person who is speaking. $H_{\mathrm{aud}}$ employs a recent breakthrough on audio localization~\cite{owens2018audio}, where the authors train a deep network to learn multisensory representation using the fact that visual and audio signals often align temporally. $H_{\mathrm{face}}$ is a machine-learning-based face detector from dlib\footnote{\url{http://dlib.net/}}. As illustrated in Figure~\ref{fig:laaf-all}D and 04:27 in the video, without LAAF the focus is always on the front person, blurring out the person in the back even when she starts talking (Row 2). LAAF analyzes future frames to understand who's speaking, and is able to correctly shift focus a few frames before the person starts to talk (Row 4).

\paragraph{New Focus Targets Evaluation}
To evaluate the set of New Focus Targets (x,y,t) generated from scene-specific LAAF, we compute the difference on focus plane in disparity units $\Delta{d}=\mathbb{D}(x,y) - \mathbb{D}(\hat{x},\hat{y})$, and the temporal position offset in number of frames $\Delta{t}=t - \hat{t}$, where $(\hat{x}, \hat{y}, \hat{t})$ is the ground truth annotated New Focus Targets, using RVR-LAAF GUI.

\subsection{Data-driven Video Autofocus \\
--- CNN-Based Selection of New Focus Targets}
\label{subsec:laaf-generic-vid}
To make autofocus fully automated on videos with any scene content, we present a first attempt at training a CNN (AF-Net) to predict New Focus Targets, replacing the scene-dependent detectors used in scene-specific LAAF. This is enabled using our RVR-LAAF GUI to annotate New Focus Targets on a large-scale video dataset. AF-Net takes in a sequence of frames centered at the query frame to predict the focus region $(x,y)$ and the probability of the query frame being a New Focus Target.

\begin{SCfigure}
  \caption{AF-Net architecture predicts the focus region ($M$) for $i_1, ..., i_{s}$ and the probability ($P$) of the center frame $i_c$ being a New Focus Target. The key of AF-Net is to cover a wide temporal range such that it sees past and future frames to make the prediction. In this illustration, CNN-1 has a temporal receptive field ($f_T$) of $2T$ and CNN-2 has a wide temporal receptive field of $2T \cdot s$, where $s$ is the number of temporal units ($f_s$).}
  \includegraphics[width=0.2\textwidth]{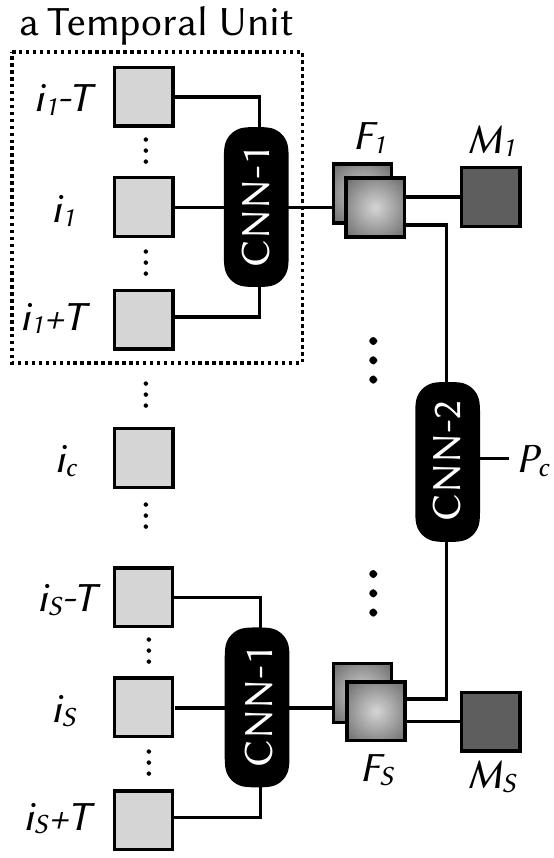}
  \label{fig:afnet}
\vspace{-5mm}
\end{SCfigure}
The key to LAAF is to analyze past and, particularly, future frames. A major challenge that arises is to have the network cover a wide temporal span of frames in a manner that is efficient in memory and computation. We introduce a temporal aggregation architecture consisting of two CNNs with different temporal receptive field sizes, as illustrated in Figure~\ref{fig:afnet}. The temporal receptive field is determined by the number of temporal units $f_s$, and the temporal coverage $f_T$ inside each temporal unit. CNN-1 has a temporal receptive field of $f_T$ to predict the focus region of the middle query frames ($i_{1}$, ..., $i_{s}$). CNN-2 takes in feature maps generated from CNN-1, and predicts the probability of the global center frame ($i_{c}$), efficiently seeing a wider temporal receptive field of $f_s \cdot f_T$ frames. Loss of the network is a weighted sum of the evaluation metrics $\Delta{d}$ and $\Delta{t}$ we described in Section~\ref{subsec:laaf-ai}.

However, evaluating $\Delta{d}$ requires disparity maps for all videos. We find that large-scale public video datasets that contains deep DOF are heavily-compressed ~\cite{abu2016youtube,wang2018revisiting} and cannot be processed by our RGBD-HDR or have tracking be applied with high fidelity. We thus made 2 adjustments to training our AF-Net. First, instead of using $\Delta{d}=\mathbb{D}(x,y)$, we use its proxy $(x,y)$ as supervision. Second, we choose to use a eye-tracking dataset DHF1K~\cite{wang2018revisiting} and a filtered version of its ground truth eye fixation map to supervise $(x,y)$, which we find to be highly correlated with focus region we annotated at New Focus Targets; supervision for $t$ comes from our annotation. We call our annotated focus video dataset VAF.

We remove unsuitable videos (\eg with jump cuts) and end up with 419 videos (640 $\times$ 360) of 20-60 seconds each in full frame rate (30 FPS). We reduce temporal sampling rate by a factor of 5 and set temporal stride of AF-Net to be 5. Both CNN-1 and CNN-2 architecture resembles ResNet-10, followed by a global average pooling for CNN-2. Training schedules are set to be the same as training our RGBD-HDR net.

\label{sec:laaf}
\section{Results}
We evaluate the key components of RVR (Section~\ref{subsec:result-rvr}), and follow the metric described in Section~\ref{subsec:laaf-action-vid} to quantitatively evaluate LAAF (Section~\ref{subsec:result-laaf}). We also show a comparison against the autofocus system inside a high-end consumer camera Olympus EM1.2 in Section~\ref{subsec:result-market}. Aside from result figures, we would like to refer readers to the accompanying video for more visually distinguishable comparisons and results.
\vspace{-1mm}
\subsection{RVR Evaluation}
\label{subsec:result-rvr}
We render shallow DOF video using the forward model in Equation~\ref{eq:forward} frame by frame, and compare rendering results with and without temporal coherency in Figure~\ref{fig:rvr-all}A. The focal planes are set to be the same for each comparison. We also show a forward model with and without occlusion-awareness in Figure~\ref{fig:rvr-all}D, and with and without HDR recovery in Figure~\ref{fig:rvr-all}B.

Evaluating intermediate network predictions on disparity and HDR maps against ground truth provides additional insight on rendering performance. For depth sensing, advanced range sensors such as LIDAR captures high-quality dense depth maps. Lightweight RGB-D cameras such as Intel RealSense~\cite{keselman2017intel} have achieved adequate resolution for some consumer-grade applications, but still suffer from noisy output and limited precision around object edges. It still remains a challenge to obtain accurate depth for casual videos using portable devices. A survey on RGBD camera is written by Zollh\"{o}fer~\cite{zollhofer2018state}. Current high-end smartphones such as iPhone X supports depth measurement using dual-pixels and dedicated post-processing to generate smooth, edge-preserving depth maps. A recent and relavant paper~\cite{wang2018deeplens} on monocular shallow DOF synthesis uses iPhone to construct the iPhone Depth Dataset for model training and testing. In a similar manner, we capture 50 test images using an iPhone X and extract disparity map as a proxy for ground truth to evaluate our predicted disparity map. We also apply a state-of-the-art monocular depth estimator, MegaDepth~\cite{li2018megadepth}, to these test images for comparison. We follow the quality metric proposed in~\cite{scharstein2002taxonomy} to compute the RMS (root-mean-squared) error measured in disparity units between the predicted disparity map and its ground truth. For HDR evaluation, it is even more challenging to capture ground truth HDR using existing hardware sensors. We choose to use the public HDR test dataset constructed from exposure stacks from HDRCNN~\cite{EKDMU17}, and use their metric by computing the mean square error in the log space of the predicted linear image and its ground truth.

\begin{figure}
\includegraphics[width=\linewidth]{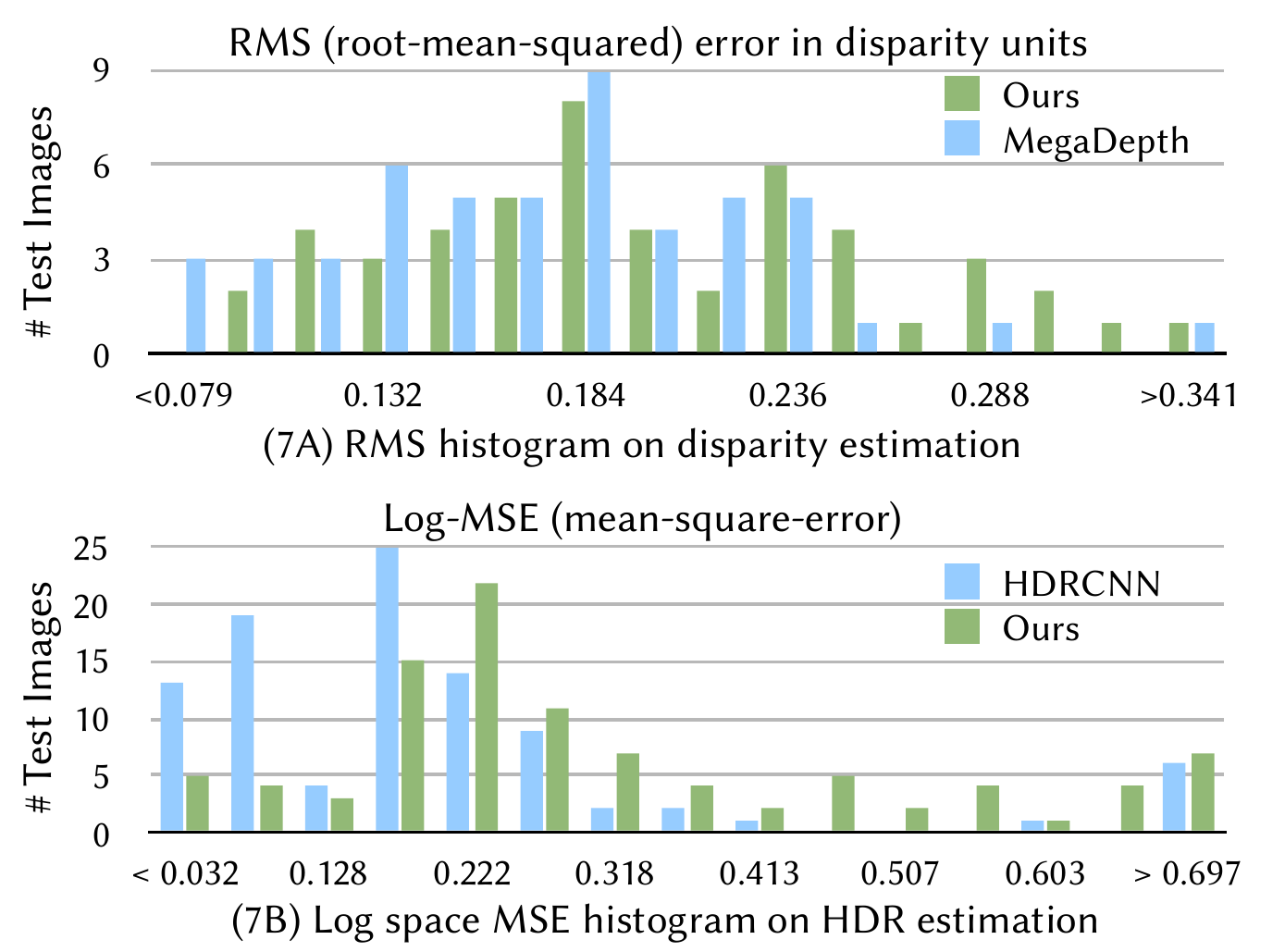}
\caption{Evaluation on predicted disparity and HDR against ground truth. Depth ground truth is obtained using dual-pixel on an iPhone X. HDR ground truth is from the public dataset described in~\cite{EKDMU17}.}
\label{fig:quant_stills}
\end{figure}

Figure~\ref{fig:quant_stills}A shows a histogram on disparity evaluation between our prediction and that from MegaDepth. Our indirect method without depth supervision produces comparable performance with MegaDepth that requires ground truth depth for training. In Figure~\ref{fig:quant_stills}B, we plot the histogram on HDR evaluation of the test images from~\cite{EKDMU17}. It is expected that HDRCNN generates better quantitative performance as the model is trained with ground truth supervision, while our model is trained without direction supervision on HDR and on a different dataset. We provide an alternative way to recover HDR without ground truth required, and are able to produce HDR maps with adequate quality to render shallow DOF images (see Figure~\ref{fig:rvr-all}B). Importantly, per frame estimation of depth and HDR is not sufficient for rendering refocusable video, as we have shown that temporal coherence is also critical in Figure~\ref{fig:rvr-all}A and video at 01:46.

%In Figure~\ref{fig:quant_stills}D, we zoom in two patches to compare between our predicted and the ground truth HDR. The blue patch shows detailed texture recovery from our prediction, and the green patch shows an challenging case of color saturation, where our predicted HDR partially recovers the original color of the flower petal.

We design RVR as a flexible system to incorporate future works that improve upon disparity and HDR estimation for still images and even for videos, or use camera sensors that support depth and HDR video streaming. A recent work~\cite{wadhwa2018} uses dual-pixel imagery to estimate scene disparity for smartphone photography. When dual-pixel imagery is available, RVR could use its depth estimation as $\mathbb{D}$ in the pipeline.

\begin{figure*}
\includegraphics[width=\linewidth]{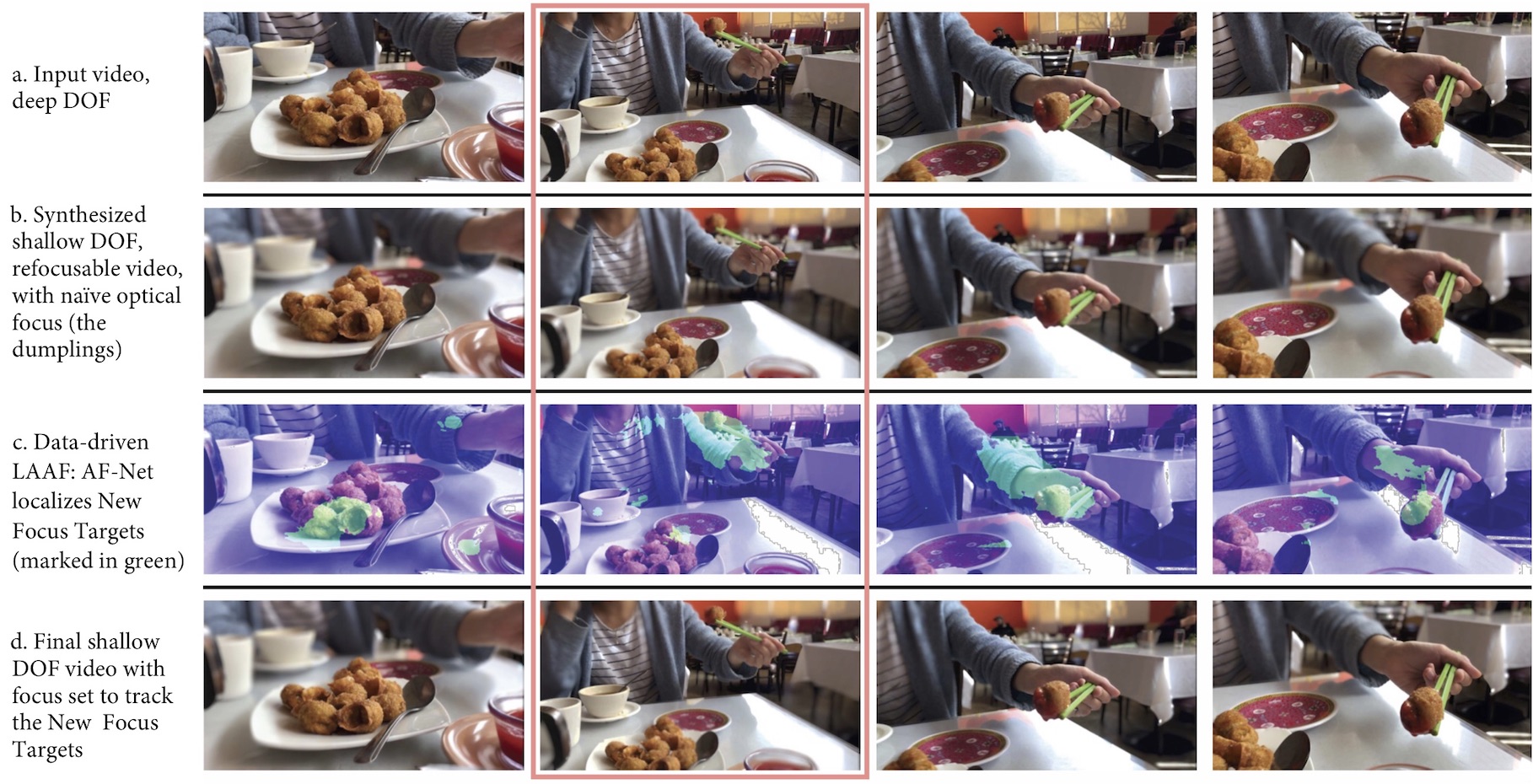}
\caption{
Data-driven AF-Net to predict New Focus Targets for video autofocus. (Row 1) Input video shows a dumpling is picked up from the plate. (Row 2) na\"ive synthetic SDOF video would easily suffer from mis-focus when applied shallow DOF. (Row 3) Focus region prediction from AF-Net; green marks focus with higher probability. Second frame (bounded by red) is the predicted temporal position of the New Focus Target, when the person picks up the dumpling. (Row 4) The resulting video shows autofocus that tracks accurately on the dumpling. Video result can be found in the accompanying video.}
\label{fig:laaf-cnn}
\end{figure*}

\subsection{LAAF Evaluation}
\label{subsec:result-laaf}
We evaluate the predicted New Focus Target $(x, y, t)$ against ground truth $(\hat{x}, \hat{y}, \hat{t})$ (from GUI annotation) using $\Delta{d}$ and $\Delta{t}$ (See Section~\ref{subsec:laaf-ai}). For each test video, we compute the average focus depth difference $\overline{|\Delta d|}$ across all frames, and the average temporal position difference $\overline{|\Delta t|}$ across all New Focus Targets.

\paragraph{GUI-based Semi-Autofocus}
We show an example in Figure~\ref{fig:laaf-all}C using RVR-LAAF GUI to annotate a New Focus Target (on the person's face). Figure~\ref{fig:laaf-all}C Row 3 presents that the user only needs to annotate the New Focus Target --- selecting a focus region and creating a tracker, and the GUI will then track the selected region. The GUI also features fine tuning on the defocus strength and the focus puller duration to account for different story tone and visual sensitivity. High-res version of the GUI is shown in the accompanying video.

\paragraph{Evaluation on Action-aware Autofocus}
We test LAAF on 11 casually collected videos with unexpected action that triggers focus depth change. We use action-aware LAAF with $\mathcal{H}=\{H_{\mathrm{sal}}, H_{\mathrm{act}}\}$, as described in Section~\ref{subsec:laaf-action-vid} to compute New Focus Targets. Among all 11 test videos, 8 (72\%) videos achieve $|\Delta t| < 15$, generating New Focus Targets that on average offset by less than half a second. Most test videos use LAAF to locate focus regions within 2 depth planes difference from ground truth. Quantitative results are shown in Table~\ref{tab:results-fm}. A qualitative result is shown in Figure~\ref{fig:laaf-all}E, and more results at 05:00 in the video.
\begin{table}[]
    \centering
    \caption{New Focus Target evaluation of action-aware LAAF on 11 casually collected videos using metrics introduced in Section~\ref{subsec:laaf-ai}. $\overline{|\Delta d|}$ (in disparity units) evaluates the performance of computed depth plane over all frames for each video. $\overline{|\Delta t|}$ (in number of frames) evaluates the average performance of temporal position of all New Focus Targets. Ground truth is obtained from hand-annotation using our RVR-LAAF GUI.}
    \resizebox{\linewidth}{!}{
    \begin{tabular}{l|c|c|c|c|c|c|c|c|c|c|c}
    \toprule
         Video ID & 1 & 2 & 3 & 4 & 5 & 6 & 7 & 8 & 9 & 10 & 11 \\
         \midrule
         $\overline{|\Delta d|}$ & 1.3 & 1.4 & 1.2 & 2.2 & 2.0 & 1.4 & 1.6 & 1.2 & 1.8 & 4.7 & 3.1 \\
         $\overline{|\Delta t|}$ & 6 & 8 & 3 & 15 & 7 & 10 & 14 & 20 & 7 & $> 30$ & 16 \\
         \bottomrule
    \end{tabular}
    }
    \vspace{.1in}
    \label{tab:results-fm}
\vspace{-5mm}
\end{table}

\paragraph{Evaluation on Data-driven Autofocus Detector}
Among the 419 VAF videos, 40 videos are held out as test set for evaluation. To evaluate New Focus Target, we run inference on a 100-frame clip to get the probability of each frame being the New Focus Target and select the frame with the highest probability. Note that for a test clip that does not have New Focus Targets (no focus depth changes required), we only evaluate $\Delta{d}$ but not $\Delta{t}$. We also test on 30 collected smartphone videos, among which 11 are the action videos used to evaluate action-aware LAAF in Table~\ref{tab:results-fm}.  $\overline{|\Delta t|}$ on the 11 action videos increase from $15.1$ (using our specialized action-aware LAAF) to $21$ (using AF-Net). However, AF-Net has the advantage of handling generic video contents and is able to achieve $\overline{|\Delta t|}$ less than 1 second on all test videos (see Table~\ref{tab:results-afnet}). This indicates the potential of having a large-scale annotated video dataset with temporal aggregation (AF-Net) to tackle the challenging problem of video autofocus. One example result is shown in Figure~\ref{fig:laaf-cnn}. Video results are at 05:58 in the accompanying video.

\begin{table}[]
    \centering
    \caption{Results using AF-Net to predict New Focus Targets on VAF test and our collected  videos. We compute the average $\overline{|\Delta t|}$ across all test videos. $\overline{|\Delta d|}$ is evaluated only on our test videos as we do not have disparity for VAF. We find AF-Net to perform slightly worse than action-aware LAAF on the action videos, but achieves reasonable performance on generic video contents, $\eg$~1 second difference in $t$ on all test videos. Ground truth for our collected test videos are from hand-annotation using RVR-LAAF GUI.}
    {
    \begin{tabular}{@{}l@{\hspace{3mm}}|c@{\hspace{3mm}}|c@{\hspace{3mm}}|c@{\hspace{3mm}}c@{}}
    % \begin{tabular}{l|c|c|c|c}
    \toprule
    Dataset & \textbf{VAF}-test & \textbf{Ours}-test &  \multicolumn{2}{c@{\hspace{0mm}}}{\hspace{6mm}\textbf{Ours}-action-test} \\
    Method &  AF-Net &  AF-Net & Action-LAAF & AF-Net \\
    \midrule
    $\overline{|\Delta d|}$ & --- & 2.64 & 1.99 & 2.21 \\
    $\overline{|\Delta t|}$ & 19.8 & 24.5 & 15.1 & 21.0 \\
    \bottomrule
    \end{tabular}
    }
    \label{tab:results-afnet}
\end{table}

\paragraph{Limitation of AF-Net}
While AF-Net explores the potential of using machine learning for video autofocus, its requirement of large-scale video focus annotation is expensive and the dataset we accumulated was of modest size. Recently, unsupervised visual feature learned from large-scale unlabeled videos has shown to be effective for video understanding, such as object tracking via video colorization~\cite{vondrick2018tracking} and audio localization by learning to temporally align audio and video~\cite{owens2018audio}. We believe video autofocus can be likewise addressed by self-supervision to learn from unlabeled internet-scale videos such as public movie clip dataset~\cite{rohrbach15cvpr}.
\vspace{-1mm}
\newcommand{\SVA}{Supplementary Video B }
\subsection{Analysis of Artifacts in Output Video}
\label{subsec:artifacts}
The performance of our RVR-LAAF system is commensurate with a first prototype according to this new approach to the video auto-focus problem. For example, visible artifacts present in rendered videos due to imperfect disparity and HDR estimation. In this section, we classify these artifacts and discuss their causes by inspection and analysis of the real video results presented in previous sections, as well as experiments on a rendered video clip (see \SVA) that provides ``ground truth" for comparison.

For the synthetic scene, we used a clip from the Blender Open Movie titled ``Sintel" (2010), rendering this scene with a simulation of a small aperture similar to the deep DOF video captured by a modern smartphone camera. In \SVA, we show ablation results that compare ground truth disparity or HDR with estimations using our method and Eilertson~\etal~\shortcite{eilertsen2017hdr} for HDR and MegaDepth~\cite{li2018megadepth} for disparity. It is important to note that there is a large domain shift between this synthetic rendering and the real training data used in all the estimation methods described, which disadvantages the estimated results. Nevertheless, the comparisons against ground truth provide empirical clues to support technical dissection of which errors in the system are associated with which classes of video artifacts.

The most visually prominent set of artifacts is due to errors in estimated disparity (see ablation comparisons in \SVA at 00:31). There are several classes of visual artifacts that may be seen. First, depth estimation across boundaries is imperfect, which is residual error in spite of our bilateral space processing. These edge errors cause prominent visual artifacts when in-focus regions are incorrectly blurred. Examples can be seen in the synthetic video result at 00:25 (ear) and in real video results at 05:13 (front person's right shoulder). The second class of visual artifact is splotchiness of synthesized defocus blur, due to incorrect spatial variation of depth estimates on flat regions. This error tends to appear in regions at large disparity, and is residual error in spite of our triplet training procedure (Section~\ref{subsec:syn-supervise}) that helps to effectively reduce this problem. Examples can be seen in the synthetic video result at 00:31 (in the highlight region of the background arm), and in the real video result at 06:20  (background segments are incorrectly rendered sharper). The third class of visual artifact related to disparity estimation error is temporal fluctuation of the defocus blur, typically in regions of background. Examples of this error can be seen in the background of synthetic video results around 0:14, and in the real video results at 04:27 (behind conversation) and 05:20 (behind dog). This is residual temporal error in spite of compensation by our temporal stabilization module.

A second set of artifacts is related to errors in HDR estimation. Common examples of this error manifest as missing bokeh balls (false negative) in output video, while hallucinated bokeh balls (false positive) are generally rare. For example, real video results at 02:20 fail to recover HDR specular highlights on the glistening sea surface and are therefore missing expected salient bokeh balls. Synthetic video results at 00:43 underestimate the HDR value of the background figure's arm and renders a darker defocused highlight.

\subsection{Compare against Market Camera}
\label{subsec:result-market}
High-end consumer cameras with state-of-the-art autofocus technology still suffers from mis-focus, especially during a rapid subject change that requires focus to resolve accordingly. We capture a pair of videos of the same scene with a Olympus EM1.2 under f/2.8 and a smartphone. RVR-LAAF is then applied to the video from the smartphone to render shallow DOF with autofocus using our GUI. We compare the two videos and show RVR-LAAF tracks focus accurately while the DSLR fails to transition focus when making large subject change (See video 06:23).

\begin{figure}
\includegraphics[width=\linewidth]{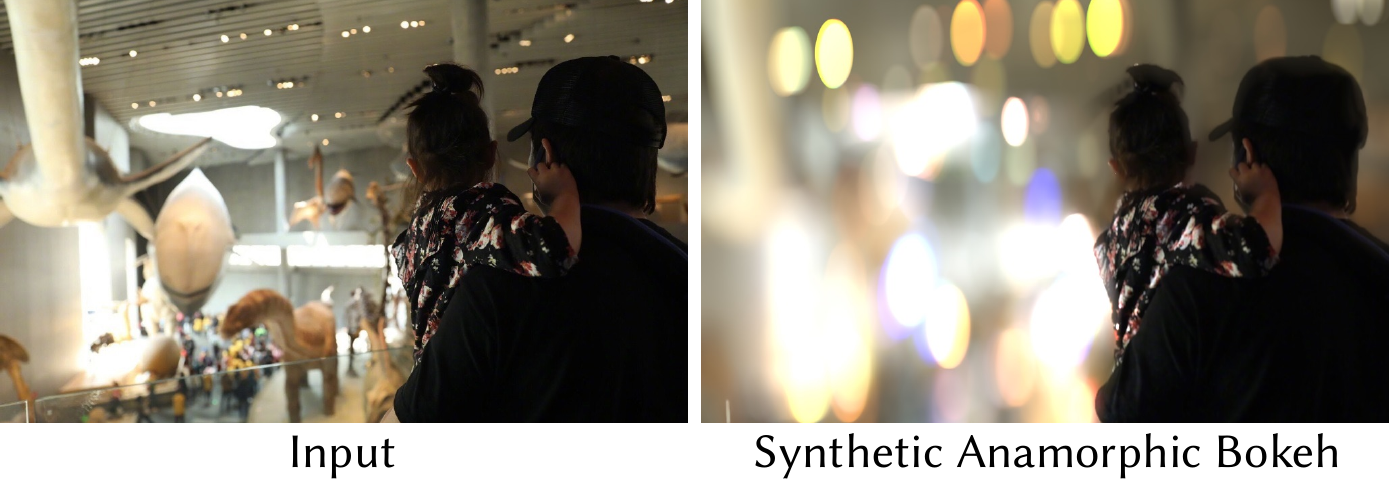}
\caption{Cinematic bokeh rendering. Our system takes in the deep DOF image (on the left), and renders a cinematic bokeh using the predicted disparity and HDR map, and a lens shape that approximates the cinematography lens ARRI Master Anamorphic.}
\label{fig:result-bokeh}
\vspace{-5mm}
\end{figure}

An interesting application is to simulate defocus produced by expensive lenses. We apply our system to demonstrate a cinematic bokeh rendering in Figure~\ref{fig:result-bokeh} that approximates the ARRI Master Anamorphic\footnote{\url{http://www.arri.com/camera/cine_lenses/prime_lenses/anamorphic/}} lens for professional cinematography. Anamorphic lenses are prized by certain cinematographers, in part because of their ellipsoidal, decidedly non-circular, defocus blur (see Figure~\ref{fig:result-bokeh}).

\label{sec:results}
\section{Conclusion}
This paper introduces the problem of delivering cinema-like focus in casual videography (\ie shallow DOF with context-aware focusing). We show that a traditional approach based on physical camera auto-focus is bound to fail, because errors in focus are baked into the video and focusing correctly in real-time requires error-prone guessing about where the action will go. We embrace this insight and take a fundamentally different approach with two parts: first, committing to rendering refocusable video from deep DOF video (RVR sub-system) rather than recording shallow DOF imagery; second,  looking at future video frames to make focus decisions at every point in the video (LAAF sub-system) rather than only looking at past frames. 

We built our RVR-LAAF prototype as a proof-of-concept for two main reasons. First, to show that we can achieve fundamentally better video auto-focus decisions by re-structuring the problem this way. And second, to show that it is tractable to attack the two sub-problems posed by this approach: synthesizing refocusable video and computing meaningful look-ahead autofocus decisions today. Regarding synthetic refocusable video, we summarized a broad array of technical approaches that the imaging and computational photography communities are actively advancing, from light field imaging to novel sensor designs to machine learning for depth inference. We are confident that performance and quality will improve rapidly. Regarding the problem of computing meaningful look-ahead auto-focus decisions, we hope to have clearly conveyed the idea that this problem is also tractable and ripe for research. We believe that this area can also advance rapidly, given the broad range of current research in computer vision that can be brought to bear.

\begin{acks}
This work is supported in part by NSF grant 1617794, an Alfred P. Sloan Foundation fellowship, and the Intel Faculty Support Program. Xuaner Zhang is supported by the J.K.Zee Fellowship Fund. We thank Alyosha Efros, Andrew Owens and anonymous reviewers for helpful discussions on the paper. We give a special thanks to people who kindly consent to be in the result videos.
\end{acks}

\bibliographystyle{ACM-Reference-Format}
\bibliography{sig2019}

\end{document}